%% file: acl_latex.tex
\documentclass[11pt]{article}

\usepackage{fancyvrb}
\usepackage{comment}
\usepackage{subfig}
\usepackage{booktabs}
\usepackage{array}
\usepackage{multicol}
\usepackage{multirow}
\usepackage{hyperref}
\usepackage[round-mode=places]{siunitx}

\usepackage{pdfpages}
\usepackage{xcolor,colortbl}
\usepackage[]{acl}
\usepackage{cleveref}
\usepackage{ragged2e}
\usepackage{tikz-qtree}

\definecolor{colorA}{HTML}{00a7d6}
\definecolor{colorB}{HTML}{d9ac36}
\definecolor{colorC}{HTML}{d04f2c}

\usepackage{times}
\usepackage{latexsym}
\usepackage{url}
\usepackage{graphicx}
\usepackage[export]{adjustbox}
\usepackage{todonotes}
\usepackage{listings}

\usepackage{textcomp}  
\usepackage{scalerel}  
\usepackage{relsize}
\def\uhel{\scalerel*{\includegraphics{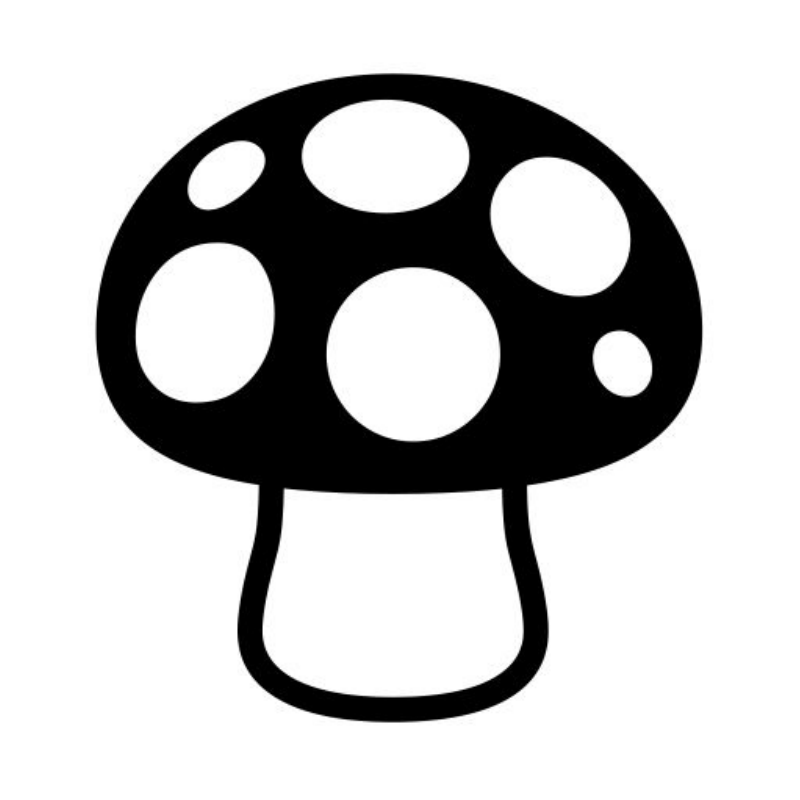}}{\textrm{\Large\textbigcircle}}}
\def\ubs{\scalerel*{\includegraphics{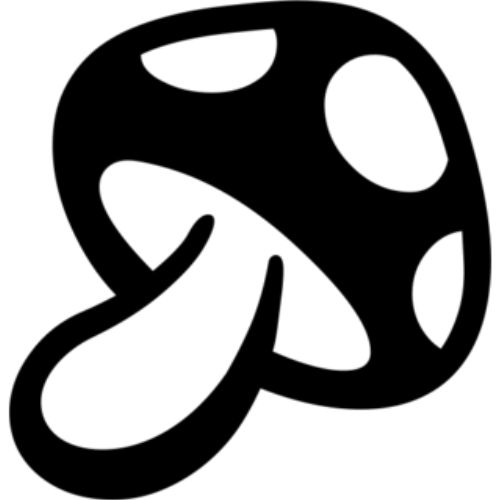}}{\textrm{\Large\textbigcircle}}}
\def\upenn{\scalerel*{\includegraphics{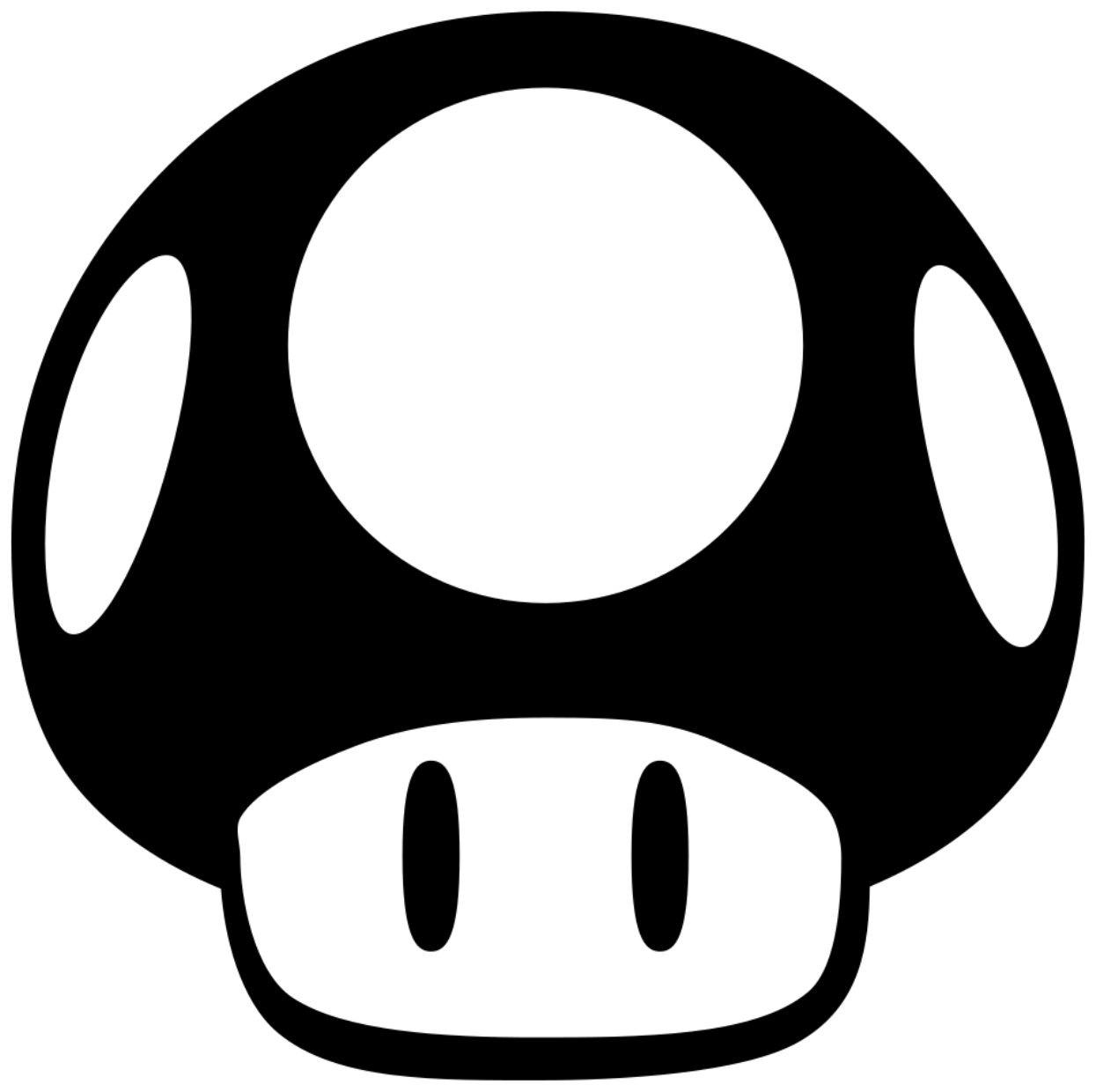}}{\textrm{\large\textbigcircle}}}
\def\silo{\scalerel*{\includegraphics{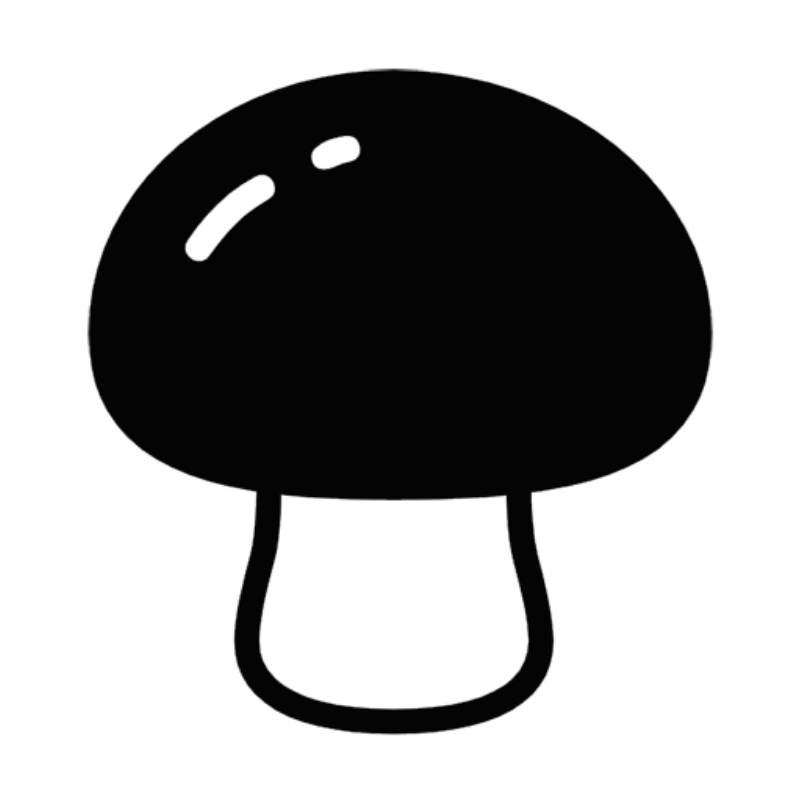}}{\textrm{\Large\textbigcircle}}}
\def\milano{\scalerel*{\includegraphics{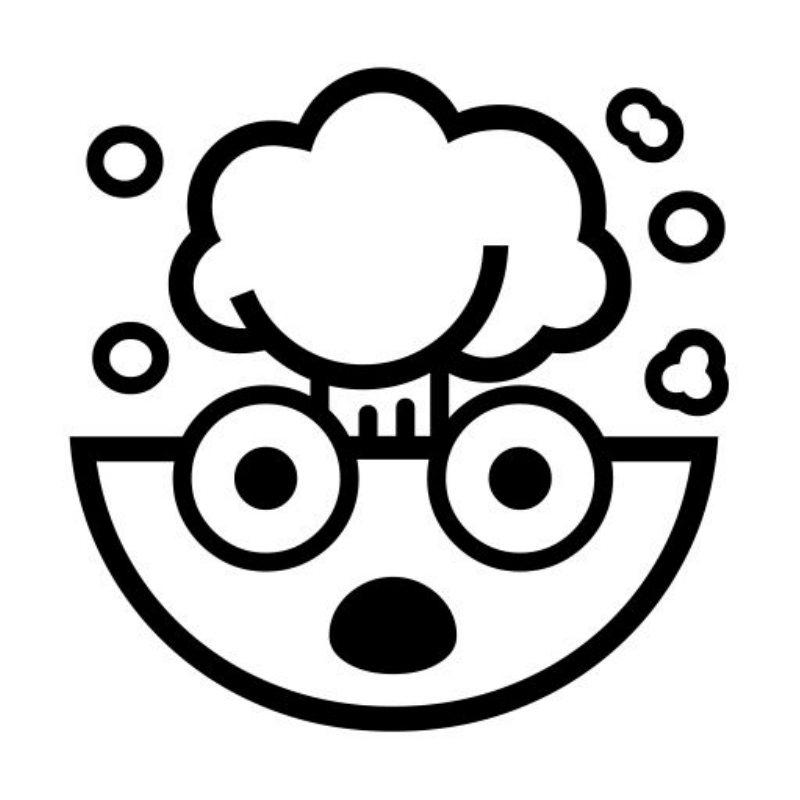}}{\textrm{\Large\textbigcircle}}}

\usepackage{fancyvrb,fvextra}
\DeclareSymbolFont{extraup}{U}{zavm}{m}{n}
\DeclareMathSymbol{\varheart}{\mathalpha}{extraup}{86}
\DeclareMathSymbol{\vardiamond}{\mathalpha}{extraup}{87}
\usepackage[T1]{fontenc}

\usepackage[utf8]{inputenc}

\usepackage{microtype}
 
\title{SemEval-2024 Task 6: SHROOM, a Shared-task on Hallucinations and Related Observable Overgeneration Mistakes}

\newsavebox{\verbbox} 

\author{
    \textbf{Timothee Mickus}\textsuperscript{\uhel} \hfill
    \textbf{Elaine Zosa}\textsuperscript{\silo} \hfill
    \textbf{Raúl Vázquez}\textsuperscript{\uhel} \hfill 
    \textbf{Teemu Vahtola}\textsuperscript{\uhel}
    \\[0.2cm]
    \textbf{Jörg Tiedemann}\textsuperscript{\uhel} \hfill 
    \textbf{Vincent Segonne}\textsuperscript{\ubs} \hfill
    \textbf{Alessandro Raganato}\textsuperscript{\milano} \hfill
    \textbf{Marianna Apidianaki}\textsuperscript{\upenn} \\[0.2cm]
    \textsuperscript{\uhel} University of Helsinki \qquad \textsuperscript{\silo} Silo AI, Finland \qquad  \textsuperscript{\ubs} Université Bretagne Sud \\[0.1cm] 
    \textsuperscript{\milano} University of Milano-Bicocca \qquad \textsuperscript{\upenn} University of Pennsylvania  \\ 
    \small{ \texttt{\{firstname.lastname\}@\{\textsuperscript{\uhel}helsinki.fi,\textsuperscript{\silo}silo.ai,\textsuperscript{\ubs}univ-ubs.fr,\textsuperscript{\milano}unimib.it\}} } \\
    \small{\texttt{\textsuperscript{\upenn}marapi@seas.upenn.edu}}
}

\begin{document}
\maketitle

\begin{abstract}
This paper presents the results of the SHROOM, a shared task focused on detecting hallucinations: outputs from natural language generation (NLG) systems that are fluent, yet inaccurate.
Such cases of overgeneration put in jeopardy many NLG applications, where correctness is often mission-critical.
The shared task was conducted with a newly constructed dataset of 4000 model outputs labeled by 5 annotators each, spanning 3 NLP tasks: machine translation, paraphrase generation and definition modeling. 

The shared task was tackled by a total of 58 different users grouped in 42 teams, out of which 27 elected to write a system description paper; collectively, they submitted over 300 prediction sets on both tracks of the shared task.
We observe a number of key trends in how this approach was tackled---many participants rely on a handful of model, and often rely either on synthetic data for fine-tuning or zero-shot prompting strategies.
While a majority of the teams did outperform our proposed baseline system, the performances of top-scoring systems are still consistent with a random handling of the more challenging items.
\end{abstract}

\section{Introduction}
The modern NLG landscape is plagued by two interlinked problems:
On the one hand, our current neural models have a propensity to produce inaccurate but fluent outputs; on the other hand, our metrics are most apt at describing fluency, rather than 
correctness.
This leads neural networks to ``hallucinate'', i.e., produce fluent but incorrect outputs that we currently struggle to detect automatically.
For instance, \citet{dopierre-etal-2021-protaugment} report that when trying to produce a paraphrase for the input ``\textit{I am not sure where my phone is}'', they obtain the following `hallucination' behavior: ``\textit{How can I find the location of any Android mobile}''.
For many NLG applications, the correctness of an output is however mission critical. For instance, producing a plausible-sounding translation that is inconsistent with the source text puts in jeopardy the usefulness of a machine translation pipeline. 

\begin{figure}
    \centering
    \includegraphics[max width=0.9\columnwidth]{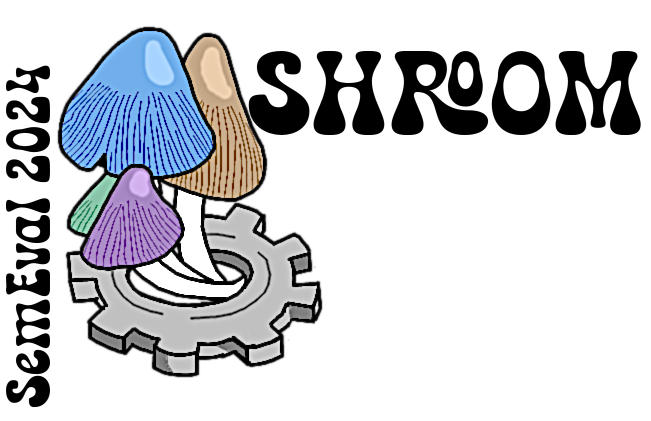}
    \caption{The SHROOM logo.}
    \label{fig:logo}
\end{figure}

This motivates us to organize a \textbf{S}hared-task on \textbf{H}allucinations and \textbf{R}elated \textbf{O}bservable \textbf{O}vergeneration \textbf{M}istakes, or SHROOM.
With our shared task, we hope to foster the growing interest in this topic in the community \citep[e.g.,][]{ji2023survey,raunak-etal-2021-curious,guerreiro-etal-2023-looking,xiao-wang-2021-hallucination,guo-etal-2022-survey}.
In particular, in the SHROOM we adopt a \textsl{post hoc} setting, where models have already been trained and outputs already produced.
Participants were asked to perform binary classification to identify cases of \textbf{fluent overgeneration hallucinations} in two different setups: \textbf{model-aware} and \textbf{model-agnostic} tracks. 
That is, participants had to detect grammatically sound outputs which contain incorrect or unsupported semantic information, inconsistent with the reference input, with or without having access to the model that produced the output. 

To that end, we constructed a dataset comprising a collection of checkpoints, inputs, references and outputs of systems covering three different NLG tasks: definition modeling \citep[DM,][]{DBLP:conf/aaai/NorasetLBD17}, machine translation (MT) and paraphrase generation (PG) trained with varying degrees of accuracy. 
Datapoints were all annotated by 5 human annotators each resulting in 1000 validation items and 3000 test items. 

Beyond simply detecting factually unsupported outputs, one of the goals of this shared task was to establish whether hallucinations are best construed as a categorical phenomenon or a gradient one.
Similar remarks have been made with respect to textual entailment \citep{bowman-etal-2015-large}.
As such, participants' submission were scored both for accuracy (whether classifiers correctly identify hallucinations) and calibration (whether classifiers are confident about their prediction when they ought to be).


The shared task attracted a total of 58 different users grouped in 42 teams, out of which 27 elected to write a system description paper.
Collectively, over the three weeks of the evaluation phase, participants submitted 300 valid sets of predictions on the model-aware track, and 320 on the model-agnostic track.
We take this participation rate, along with the breadth of methodological approaches developed by participants, as clear signs of success for our shared task: 
This large pool of participants allows us to identify and discuss some key trends in how the task was tackled.
Crucially, many participants rely on a handful of model, and often rely either on synthetic data for fine-tuning or zero-shot prompting strategies.
In terms of raw performance, we note that while a majority of the teams ($64$ to $71\%$) did outperform our proposed baseline system, the performances of top-scoring systems are still consistent with a random handling of the more challenging items.
In sum, this first iteration of the SHROOM underscores both an interest of the research community as well as the current limitations in our approaches.

The remainder of this article is structured as follows: Iin \Cref{sec:sota}, we provide an overview of the current research landscape.
\Cref{sec:def} defines our theoretical framework, and \Cref{sec:data} summarizes our data collection process.
We then present and discuss shared task results in \Cref{sec:perfs,sec:quali} before concluding with a few thoughts on further research in \Cref{sec:ccl}.


\section{Connecting with the past: related works and state of the art}
\label{sec:sota}

It is now widely accepted that NLG models often generate outputs that are not faithful to the given input, commonly referred to in the community as hallucinations \citep{vinyals2015neural,raunak-etal-2021-curious,maynez-etal-2020-faithfulness}. Yet there is minimal consensus on the optimal framework for its application. This lack of agreement is due in part to the diversity of tasks that NLG encompasses \citep{ji2023survey}. 

\citet{guerreiro-etal-2023-looking} propose a taxonomy of hallucinations that includes oscillatory productions, and fluent but strongly or fully ``detached'' outputs.
While this taxonomy is well constructed, we find it inadequate for the needs of the community at large for four reasons: 
\textit{(i)}~It conflates some issues of fluency with semantic correctness (oscillatory productions are cases of non-fluent overgeneration where no extraneous semantic material is introduced); 
\textit{(ii)}~It only considers the most extreme cases of hallucinations (strongly or fully detached productions), whereas diagnosis of intermediary cases is bound to be more challenging and useful to the community; 
\textit{(iii)}~It focuses only on MT, although other tasks are also known to suffer from fluent overgeneration \citep[e.g.,][]{rohrbach-etal-2018-object}, including the ones we propose to address; 
\textit{(iv)}~It uses only lowest scoring outputs, whereas any tool built to verify system outputs ought not to flag non-pathological outputs. 

Alternative studies have built benchmarks for hallucination detection, with a predominant emphasis on dialogue systems.
\citet{li-etal-2023-halueval} propose the HaluEval benchmark using an annotation framework that does not necessarily center on the input given to the model and requires the annotators to search the internet for facts. Moreover, they opted to annotate the outputs of a popular LLM, with the major downsides that it is closed, not-transparent and commercial; rendering the research outputs that may stem from future studies less interesting.
Other benchmarks include the works of \citet{liu-etal-2022-token} and \citet{zhou-etal-2021-detecting}, which automatically insert hallucinations into training instances to generate syntactic data for token-level hallucination detection; \citet{lin-etal-2022-truthfulqa}, which work with factual claims supported by reliable, publicly available evidence; and \citet{dziri-etal-2022-evaluating}, which focus on knowledge-based dialogue systems and base their annotation on NLI, relying only on the system's input, just as we do.

\section{Tripping over hallucinations: task definition and annotation}
\label{sec:def}
In contrast with previous works \citep[e.g.][]{guerreiro-etal-2023-looking, li-etal-2023-halueval}, we focus on cases of fluent overgeneration since judgments pertaining to the over-generative nature of a production can be elicited by means of \textbf{inferential semantics}: if an output cannot be inferred from its semantic reference, then it contains some information that is not present in the reference---i.e., the model has generated more than we expected.\footnote{
    Note that if the output can be inferred from the reference but the information is not explicitly present in the reference, then the model is actually making a correct semantic inference: it is generating a semantically sound output. 
    E.g., if the the model produces ``my tie is blue'' for the reference ``my tie is the color of the sky'', the model output is semantically sound.
}
This approach connects with the theoretical framework sketched by
\citet{vandeemter2024pitfalls}, who likewise relies on inferential semantics but also considers undergeneration issues in NLG outputs.
We provide multiple annotations and a gold majority label, given the low consensus on semantic annotations \citep{nie-etal-2020-learn}.

\begin{figure}
    \centering
    \Tree [ [ \textcolor{colorA}{DM} \textcolor{colorC}{MT} \textcolor{colorC}{PG} ].\emph{Model-agnostic} [ \textcolor{colorA}{DM} \textcolor{colorC}{MT} \textcolor{colorC}{PG}  ].\emph{Model-aware} ] .SHROOM
    \caption{Shared task overview. Both tracks feature all three NLG tasks. Datapoints from systems in \textcolor{colorA}{blue} correspond to target-referential datapoints and  in \textcolor{colorC}{red} the ones that are either target- or source-referential; which we refer to as \textit{dual-referential}.
    }
    \label{fig:overview}
\end{figure}

In \Cref{fig:overview} we provide an overview of the task.
The SHROOM is framed around two key distinctions: (i) model-aware vs. model-agnostic approaches, and (ii) source-referential vs. dual-referential datapoints.
The former corresponds to whether participants have access to the model that generated the item: 
\textbf{Model-agnostic} approaches are practical, as models may not be accessible to end users; 
\textbf{Model-aware} approaches can lead to richer and more accurate diagnoses. 
The latter is a consequence of our inferential take on over-generation: what can effectively serve as a semantic reference varies across NLP systems.
For DM, where we fine-tune a language model to produce a definition for a given example of usage the datapoints are \textbf{target-referential}, i.e. the target is the sole usable semantic reference. In this context, the target serves as the sole usable semantic reference. Conversely, the target is expected to be semantically implied from the source in source-referential tasks, such as summarization. Note that we do not annotate source-referential tasks due to annotation challenges that make them unreliable for our purposes. In dual-referential tasks like PG \& MT, this distinction bears no weight.

  \begin{figure}
    \centering
    \begin{lstlisting}[basicstyle=\fontsize{7.5}{9.5}\ttfamily]
{ "hyp":"A cigarette .",
  "ref":"tgt",
  "src":"I stepped outside to smoke myself a j . 
         What is the meaning of J ?",
  "tgt":"( plural Js or J 's ) A marijuana
          cigarette .",
  "model":"ltg\/flan-t5-definition-en-base",
  "task":"DM",
  "labels":["Hallucination","Not Hallucination",
           "Not Hallucination", "Hallucination", 
           "Hallucination"],
  "label":"Hallucination",
  "p(Hallucination)":0.6  }
\end{lstlisting}
    \caption{Target-referential datapoint example from the validation set for the model-aware track.}
    \label{fig:example} 
    
\end{figure}

In ~\Cref{fig:example}, we present an example datapoint displaying how we plan to encode all relevant information in a JSON format is provided.
The datapoint keeps track of the source provided to the model as input (\texttt{src}), the intended target (\texttt{tgt}), the model production (\texttt{hyp}), the task this production was derived from (\texttt{task}), can correspond to DM, MT or PG), whether this datapoint is target-referential (\texttt{ref}), the annotations, the gold label and the proportion of annotators that labeled the utterance as a hallucination (\texttt{labels},  \texttt{label}, and \texttt{p(Hallucination)}).
In the model-aware track, we will also provide a HuggingFace model name (\texttt{model}).

\section{Foraging and harvesting season: Collected data}
\label{sec:data}

All SHROOM data (models, outputs and annotations) are available under a CC-BY license.\footnote{See \href{https://helsinki-nlp.github.io/shroom/}{\tt helsinki-nlp.github.io/shroom}}

\subsection{Data \& model provenance} 
Participants have access to generated outputs from multiple systems trained to generate English output at various stages of their training, stemming from three sequence-to-sequence NLG tasks: DM, MT and PG. 
The SHROOM dataset consists of annotated \textit{test} and \textit{dev} sets, as well as a \textit{unlabeled training split} of 30k datapoints per track  and the full set of possible target references to allow corpus-wide approaches. To ensure effective annotation of the development and test sets, and to be able to guarantee a gradient in quality as measured by automated metrics, we pre-selected fluent outputs for the annotators, which we describe in the following.\footnote{Note that we do not warranty that the training split contains fluent outputs.}

\paragraph{MT:} For the model agnostic track we use the models from \citet{mickus-vazquez-2023-bother}. We compute perplexity for the all MT outputs and BERTScores with regards to the outputs and corresponding targets. We filter outputs with perplexity scores above the 2\% quantile. From the filtered outputs, we randomly select 200 samples with BERTscores in the 1/7, 2/7, 3/7, 4/7, and 5/7 quantiles.  For the model-aware track, we use the NLLB model \citep{nllbteam2022language} and produce translations on the Flores-200 dataset from languages marked as low-resource to English. Next, we manually select a sample that is sufficiently fluent.

\paragraph{DM:} We use the model of \citet{segonne-mickus-2023-definition} for the model-agnostic track, and for the model-aware track we used the \texttt{flan-t5-definition-en-base} \citep{giulianelli-etal-2023-interpretable}. We generate outputs on the English portion of the CoDWoE dataset \citep{mickus-etal-2022-semeval}, and manually select a sample that is reasonably fluent and contains no profanities.

\paragraph{PG:} We used a pretrained and fine-tuned paraphrasing model\footnote{\url{https://huggingface.co/tuner007/pegasus_paraphrase}} based on Pegasus \citep{zhang2020pegasus} for the model-aware track, and the controlled paraphrase generation model of \citet{vahtola-etal-2023-guiding} for the model-agnostic track.

We generated paraphrase hypotheses using Europarl \citep{koehn-2005-europarl} and Opusparcus \citep{creutz-2018-open} for the model-aware and -agnostic tracks, respectively. 
For the model-aware setup, we generated 50 hypotheses for each source sentence using diverse beam search \citep{vijayakumar2016diverse} using BLEU scores \citep{papineni-etal-2002-bleu} to select the least similar hypothesis for each source sentence to serve as its paraphrase.
For the model-agnostic setup, we calculated control tokens for each source sentence as in \citet{vahtola-etal-2023-guiding}, scaled the length-controlling value in range (1, 1.5) with a uniform probability distribution to provoke hallucination in the generated sequences, and used beam search with a beam size of 5 to produce the paraphrases.
We manually curated the final validation and test examples. 

\subsection{Annotation}
We annotate a total of 4,000 items, which are split 25\%--75\% between development and test sets: 1000 datapoints come from PG, 1500 from DM and 1500 from MT. Each item is annotated by five annotators on whether the reference entails the output. Annotations are binary, for ease of dataset construction. 
Gold labels are defined with respect to the annotators' majority vote.

The annotators were enlisted via Prolific,\footnote{\url{https://www.prolific.com/}} a paid platform specialized in gathering human data for research studies and AI dataset creation, among other purposes. 
We did not target any particular group of participants; the only screening prerequisites were that (i) participants had to be fluent in English and (ii) they should not have taken part in an initial pilot study.

We used \href{https://github.com/davidjurgens/potato}
{Potato} \citep{pei-etal-2022-potato}, an open-source annotation tool specifically designed to seamlessly integrate with Prolific. 
Annotators were first presented with a pre-annotation screen outlining the annotation guidelines, after which they commenced the annotation of items individually. 
Each item consisted of the Reference, the AI-generated output, and relevant context regarding the NLG task (DM, MT, or PG).  
The annotators were asked to answer the question \textit{"Does the following AI output only contain information supported by the Reference?"} responding with either "yes" or "no," and were also given the opportunity to provide comments if necessary. 
Additionally, they could navigate back and forth through their assigned items. We set up a timer that notified the participants every 60 seconds of the time spent on an item.
In \Cref{appx:annotation_guidelies}, we present a copy of the instructions we used.

To control for annotation quality, we manually reviewed annotations from two sets of selected annotators: (i) five randomly selected annotators; and (ii) the five annotators who completed the task the fastest (under 3.5 minutes). 
All 10 annotators completed 20 annotations each. 
We judged all 200 annotations to be sound, in that a reasoning could be reconstructed to explain the provided annotation.

\begin{figure*}[th]
    \centering
    \subfloat[\small{Validation split}]{
        \includegraphics[max width=0.475\linewidth]{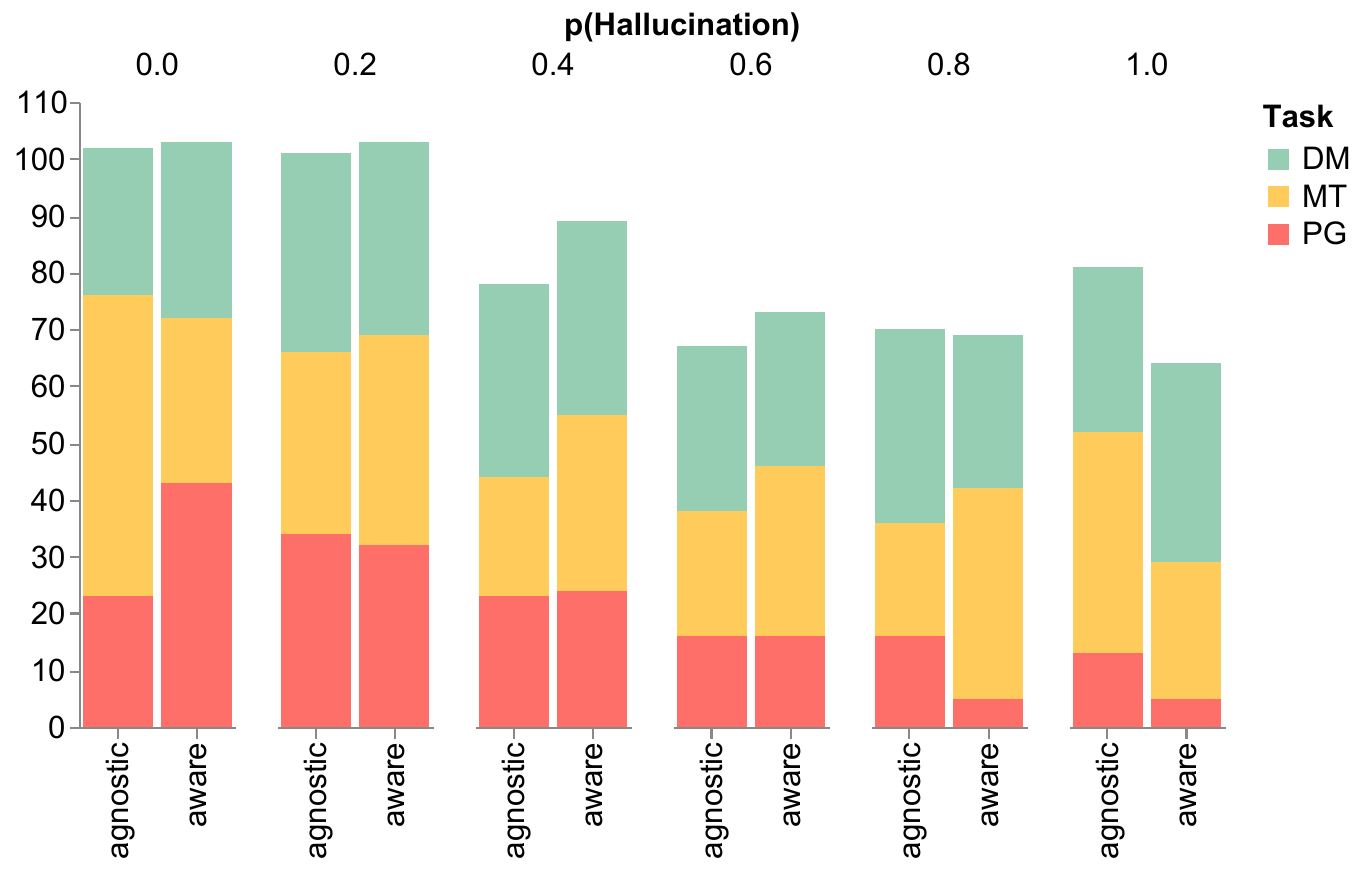}
    }
    \subfloat[\small{Test split}]{
        \includegraphics[max width=0.475\linewidth]{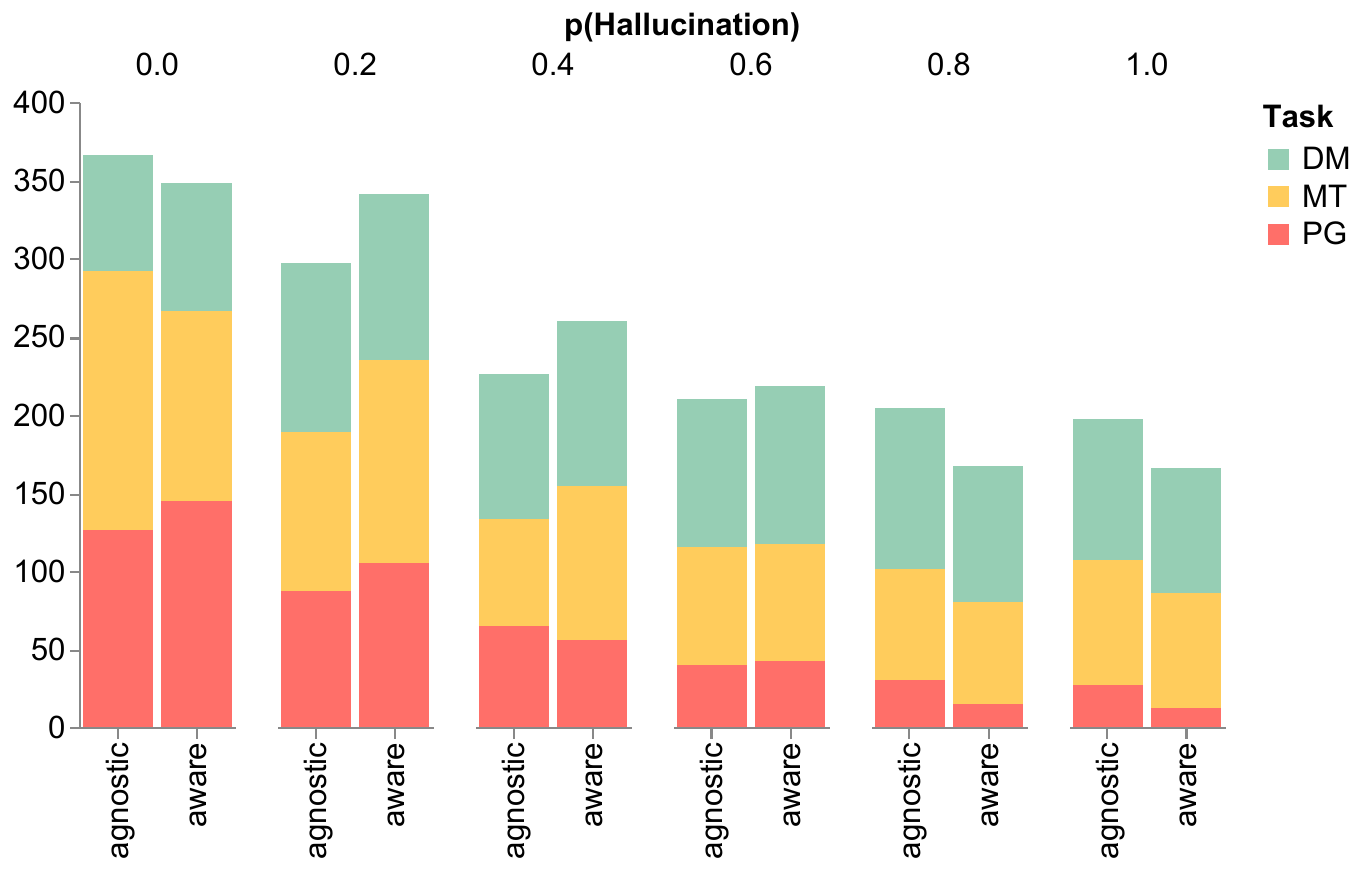}
    }
    \caption{Distributions of annotations   
}
    \label{fig:label-dist}
\end{figure*}

\paragraph{Label distribution.} \Cref{fig:label-dist} provides an overview of the distribution of labels in the SHROOM dataset splits (validation and test), broken down per NLG task (MT, DM and PG) and track (model-aware vs. model-agnostic).
In this figure, we consider the empirical probability that a given item is judged to be a hallucination, i.e., the proportion of annotators judging the NLG output is not supported by the intended semantic reference.

We can highlight two trends in this figure.
The first one, and perhaps most important, is that hallucinations are not consensual among our annotators.
If intuitions regarding hallucinations were clear-cut, we would strongly expect a bi-modal distribution of empirical label distributions being consistently judged as hallucinations or not hallucinations. 
Instead, we find a number of intermediate cases, where annotators are split: These account for $29$--$32\%$ of the data, depending on the split (validation or test) and track (model-aware or model-agnostic).
Given the small number of annotators per datapoint, we cannot confidently rule out the possibility of a sampling bias---it is plausible that a larger pool of annotator would yield more bimodal empirical distributions.
On the other hand, this tentative evidence is also in line with what has been argued elsewhere for natural language inference \citep{nie-etal-2020-learn,zhou-etal-2022-distributed}.
This is in fact well exemplified by the datapoint provided in \Cref{fig:example}: Whether the term \emph{cigarette} is underspecified and can apply to any smokable substance, or whether it is to be understood as prototypically referring to {tobacco cigarettes} by default is, in fact, up for discussion---and it stands to reason that different speakers may form different opinions.

Second, it is difficult to find hallucinations: The higher the empirical probability, the fewer the datapoints.
This is especially true in the PG task: these outputs rarely yields consensual hallucinations, whereas we can find such items in DM and MT much more frequently.
Looking at the expected value of the empirical probability per task, we find that DM consistently ranks higher than MT, which in turns ranks higher than PG.
Both of these differences are significant under a one-sided Mann-Whitney U-test in the two test tracks ($p < 0.0003$); in the model-aware validation dataset, only the difference between MT and PG is significant ($p < 2 \cdot 10^{-8}$), in the model-agnostic validation dataset, only the difference between DM and MT is ($p < 0.04$).
We note that DM requires a more complex processing of its input, as it has to rely on facts captured by the underlying LLM during its pre-training phase; for MT and PG, the input of the NLG task contains the semantic information necessary to produce a valid output.
As such, we conjecture that the difficulty of an NLG task fosters hallucinatory behavior.\footnote{
    We also remark that the two tracks are broadly comparable in terms of hallucinatory content.
    Two-samples Kolmogorov-Smirnov tests for either split (test or validation) do not provide sufficient grounds to suggest a difference of distribution in labels between model-aware and model-agnostic tracks---which again suggests that the relevant difference is at the task level, rather than at the model level. 
}

\section{They got so high: shared task results}
\label{sec:perfs}

The competition was held via Codalab \citep{codalab_competitions_JMLR}.
The leaderboard was left hidden during the evaluation phase (i.e., participants were not notified of their submissions' scores until the end of the evaluation phase) but users were allowed to make a high number of submissions (50). 

Systems are evaluated according to two criteria: the \textbf{accuracy} that the system reached on the binary classification, and their \textbf{calibration}, measured as the Spearman correlation of the systems' output probabilities with the proportion of the annotators marking the item as overgenerating.
We rank systems by accuracy and break possible ties using calibration.

\subsection{Baseline system}
As a baseline for the task, we use an LLM\footnote{We use quantized \texttt{Mistral-7B-Instruct-v0.2}~\cite{jiang2023mistral}, from the \texttt{Hugging Face hub} \url{huggingface.co/TheBloke/Mistral-7B-Instruct-v0.2-GGUF} or the \texttt{llamacpp} project {\url{github.com/ggerganov/llama.cpp}}.} to evaluate whether the generated hypotheses are coherent with the provided context. Drawing upon \citet{manakul-etal-2023-selfcheckgpt}, we use the prompt template listed in \Cref{fig:prompt}.
The system of \citet{manakul-etal-2023-selfcheckgpt}, which has gathered some attention from the community, constitutes a straightforward approach based on a modern LLM, and is therefore well-suited to serve as a baseline in our shared-task: it corresponds to a reasonable default approach to tackle the problem we challenge participants with.

\begin{figure}
    \centering
    \resizebox{\columnwidth}{!}{\parbox{1.175\columnwidth}{%
\small
\texttt{Context: \{\}}\\
\texttt{Sentence: \{\}}\\
\texttt{Is the Sentence supported by the Context above?}\\
\texttt{Answer using ONLY yes or no:}
    }}
    \caption{Prompt template used in the baseline system, adapted from \citet{manakul-etal-2023-selfcheckgpt}.}
    \label{fig:prompt}
\end{figure}

The specific context varies depending on the task addressed, i.e. the source sentence for the paraphrase generation task, and the target sentence for machine translation and definition modeling tasks. As for the probability of hallucination, we rely on the probability assigned by the model to the first output word.\footnote{We note that this simple heuristic may not accurately represent the true hallucination probability.} In cases where the output does not clearly indicate \textit{yes} or \textit{no}, we randomly select one, attributing a hallucination probability of $0.5$. 

On the model-agnostic track, our baseline system achieves an accuracy of $0.697$ (with a calibration of $\rho=0.403$), on the model-aware track, we observe an accuracy of $0.745$ (with $\rho=0.488$).
We can also indicate some other simple heuristics, such as picking the most frequent label (viz., \texttt{Not Hallucination}):
In this case, one would expect an accuracy of $0.593$ on the model-agnostic track, and $0.633$ on the model-aware track.
A purely random guess between the two possible labels would result in an accuracy of $0.5$.
In short, our baseline systems systematically outperforms these crude heuristics. 

\subsection{Participating teams}
A total of 59 individual users grouped in 42 teams participated in the shared task, out of which 27 elected to write a system description paper. 
During the evaluation phase, we received a total of 512 submissions, out of which 368 were successful.
264 of these submissions targeted both tracks, while 68 only targeted the model-agnostic track, and 36 only targeted the model-aware track. That is, we received 332 model-agnostic submissions and 300 model-aware submissions.

\begin{table*}[!h]
\centering
    \subfloat[\label{tab:ranking_agnostic} Model-agnostic track rankings]{%
        \resizebox{0.8\width}{!}{%
        \input{tables/rankings-agnostic}%
        }%
    }%
    \subfloat[\label{tab:ranking_aware} Model-aware track rankings]{
        \resizebox{0.8\width}{!}{
            \raisebox{4.1ex}{\input{tables/rankings-aware}}
            \vphantom{\input{tables/rankings-agnostic}}
        }
    }
    \caption{SHROOM team rankings. Codalab usernames are used to define teams when no other information was provided.}
    \label{tab:ranking}
\end{table*}

We present the model-agnostic track rankings in \Cref{tab:ranking_agnostic} and the model-aware track in \Cref{tab:ranking_aware}. 
As one might expect, there is a high correlation between the accuracy and calibration scores of each team's top ranking submission, which translates into a Spearman's $\rho$ correlation of $0.909$ on the model-agnostic track and $0.949$ on the model-aware track.
Most of the top submissions per team rank above our baseline ($30/42\approx71.4\%$ in the model-agnostic track, $25/39\approx64.1\%$ in the model-aware track).
This appears roughly in line with all submissions globally: $69.9\%$ of all model-agnostic submissions and $57.0\%$ of all model-aware submissions score higher than our baseline.

Another point worth stressing is that teams that fare well on one track usually fare equally well on the other: For the 38 teams participating in both tracks, we find that the rank they obtain on the model-aware track correlates with the rank they obtain on the model-agnostic track (Spearman's $\rho=0.884$).
This would tentatively suggest that participants could not effectively leverage the supplementary data available in the model-aware track.\footnote{An alternative account would be that all teams that participated in both tracks equally benefited from the access to the model weights, which we deem much less likely.}

Lastly, we note that there is a ceiling in terms of performances: 
The most effective systems misclassify between 15 to 19\% of all items, or almost one in every six or five datapoints. 
We have discussed above that, as hallucinations are a graded phenomenon, a large segment of our data (30\%) corresponds to ambiguous cases where annotators are split 2 vs. 3.
As such, it is worth stressing that top scores are consistent with models that classify consensual items well (where at most one annotator disagree), but perform at random chance on the more challenging ambiguous datapoints.

\section{A bunch of fun guys: qualitative analysis of participants systems}
\label{sec:quali}

We derive our analyses from system description papers as well as self-reports from a handful of participants who elected to not provide a full description of their systems.
This corresponds to 33 systems out of the 42 identified teams that participated to the shared task, out of which 6 did not provide a full description.
See also \Cref{tab:all_teams} in \Cref{adx:participants} for further details.

\begin{figure}
    \centering
    \includegraphics[clip, trim=0cm 3.5cm 0cm 0cm, max width=\columnwidth]{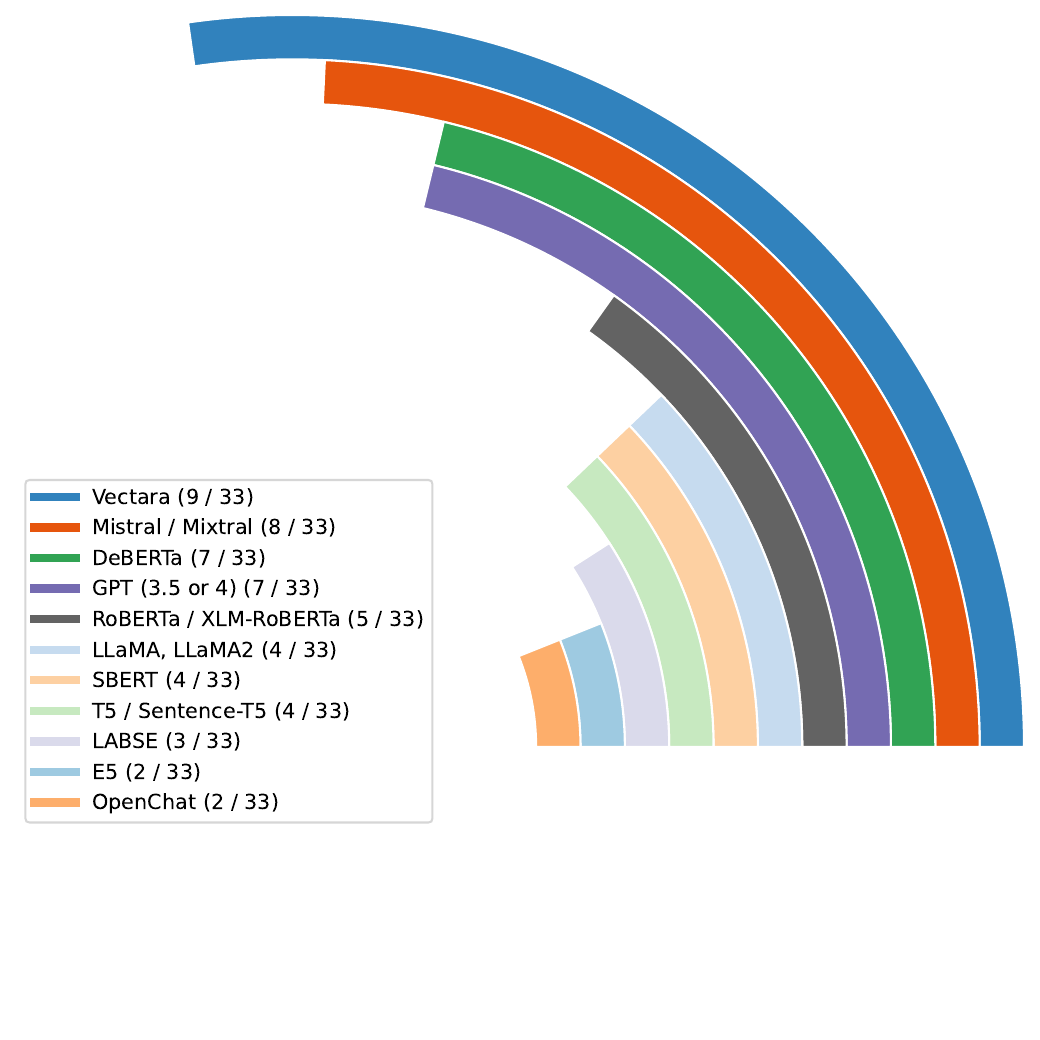}
    \caption{Known models used by more than one team. A full circle would correspond to a given model used by all of respondents, half a circle to 50\% of respondents using said model. Best viewed in color.}
    \label{fig:models}
\end{figure}
    
\paragraph{How the task was approached.} 
The teams used a variety of methods to address the problem, ranging from ensemble techniques to fine-tuning pretrained language models (LLMs) and prompt engineering. 
As expected, most teams used popular pre-trained LLMs such as GPT,  LLaMA, DeBERTa, RoBERTa, and XLM-RoBERTa; \Cref{fig:models} provides a summary of which models were most popular among our teams.
The Vectara hallucination evaluation model\footnote{\url{https://huggingface.co/vectara/hallucination_evaluation_model}} turned out to be extremely popular, as more than 1 in 4 teams that provided information about their systems report having used it in their experiments.
If we add other DeBERTa-based models, this number climbs to $16/33$, i.e. almost every other team used DeBERTa or a variant thereof.

Yet, the ways in which these LLMs were used cover a wide range of approaches: Some either fine-tuned on hallucination data or optimized with prompts; others employed in-context learning with role-playing, automatic prompt generation, and ensemble methods. Furthermore, some teams focused on zero-shot and few-shot approaches, while others focused on synthetic data generation and semi-supervised learning techniques to construct a labeled training set. Especially noteworthy, \citet{no23} report constructing a manual dataset of 3000 datapoints for training their systems.
 
Teams predominantly relied on the data constructed for the SHROOM, although some teams added datasets such as QQP and PAWS. 
Interestingly, we also note five teams relying on NLI/entailment data or models, including some that achieved high results (\citealp{no200,no237,no256,no269} and Team CentreBack)---and this matches the theoretical framework adopted in this shared task.

\paragraph{What worked well.} We now turn to what distinguishes top scorers from other submissions. We note that systems based on the closed-source models GPT-3.5 and GPT-4 tend to fare well:  4 out of the 6 highest scoring systems on either track---\citet{no53,no200,no269,no126} and Alejandro Mosquera---all report using these models.
This is however not a strict requisite as OPDAI \citep{no109} manages to rank high (2\textsuperscript{nd} on the model-agnostic track and 4\textsuperscript{th} on the model-aware track) without it.
Neither does using closed-source models guarantee a high result: UCC-NLP and \citet{no40} also use GPT-3.5, and while the former is ranked 14\textsuperscript{th} on the model-agnostic track and 13\textsuperscript{th} on the model-aware track, the latter is ranked 30\textsuperscript{th} on the model-agnostic track and 21\textsuperscript{st} on the model-aware track, and only outperforms the baseline model in accuracy by 0.02 to 0.03 points.

Remarkably, many of the top-scoring approaches rely on fine-tuning \citep{no269,no200,no182,no109} or ensembling (\citealp{no53,no43}, Alejandro Mosquera), suggesting that high performances do not come out of the box from off-the-shelf LLMs and systems.
It is necessary to adapt existing models or properly establish to what extent their predictions are useful to the task at hand.

\begin{figure}
    \centering
    \includegraphics[max width=\columnwidth]{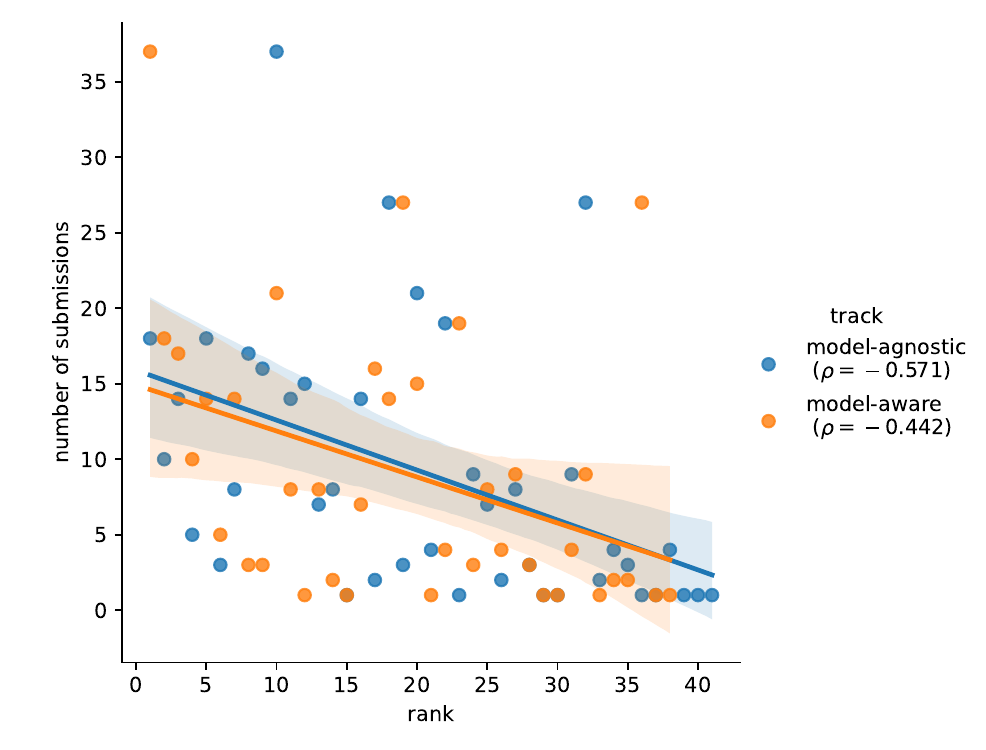}
    \caption{Rank obtained vs. number of submissions made on both tracks.}
    \label{fig:rank-vs-sub}
\end{figure}

Another important trend we identify is that the number of submissions per team anti-correlates with the rank they obtain: 
The more participants submitted, the higher their best scores went.
This is visualized in \Cref{fig:rank-vs-sub}: 
On both tracks, we find reasonable anti-correlations ($-0.58 < \rho < -0.44$) indicating that top-scorers tended to submit more.
This might provide an alternative explanation for what distinguishes top-scorers from other participants: 
If we were to model participants' submissions as a random process, we would expect that sampling more often (i.e., submitting more) would mechanically yield a better rank. 

Overall, the high methodological diversity highlights the complexity of hallucination detection, even when contained the simple inferential semantics framework of our shared task:
While a focus on NLI or using high-performance closed source models may help, the highest scores are obtained through thorough involvement---both in terms of model training and prediction set submissions.


\section{Much room to grow: conclusions and future perspectives}
\label{sec:ccl}

This first iteration of the SHROOM shared task on detecting hallucinations has allowed us to make significant headway into understanding the confabulatory behavior of modern NLG systems.
The data collected demonstrate that \emph{hallucinations correspond to a gradient phenomenon}, and that different speakers form different opinions as to what counts as a hallucination.
We were also able to showcase that \emph{ambiguous items remain challenging}, and that the current state of the art on the dataset we provided is compatible with simple random guesses whenever the data is more ambiguous. 
This results underscore the massive gap that NLP research urgently needs to address: one out of every six items is still misclassified by the most effective systems showcased during this shared task.

The diversity of methodologies employed by participants underscores how \emph{out-of-the-box solutions are not sufficient}: Highest scoring teams had to rely on fine-tuning or ensembling and made a high number of submissions.
Relatedly, \emph{access to the model parameters was of limited help}: 
Few approaches attempted to perform model-specific investigations, and performances on the model-aware track are in fact lower than what we observed on the model-agnostic track.
Properly leveraging the parameter space for finer-grained hallucination detection remains a point for future research to investigate.

This shared task has not broached some crucial aspects and questions: How do these results translate insofar as modern LLMs---often much larger and better trained than the systems we studied here---are concerned? 
Can we leverage sentence-level predictions to pinpoint token-level issues with the output of our NLG systems?
And will the difficulties that we underscored in this purely English be exacerbated when studying other languages---especially those that are less well-resourced and typologically different?
Answering these questions and more will require further research---and perhaps future iterations of this shared task.

Overall, the success of this shared task is owed to its committed participants.
We received over 350 submissions in the span of three weeks from across the world. 
The width of approaches studied and reported upon provides a useful snapshot of where the field is at, what approaches are favored, and what gaps still need to be overcome.
We expect that the results of the SHROOM will provide a useful starting point for future work on hallucinations.


\section*{Doing SHROOM responsibly: ethical considerations}

We strive to adhere to the \href{https://www.aclweb.org/portal/content/acl-code-ethics}{ACL Code of Ethics}.

\paragraph{Broader Impact.}
Hallucinated outputs from large language models can be used to further spread disinformation and advance misleading narratives. Detecting hallucinated outputs is an important step in elucidating the factors of this phenomena and contribute to ongoing efforts to mitigate hallucination. This leads to the development of more trustworthy generative language models.

\paragraph{Data and Annotators.}
Our annotators were suitably compensated for their work in excess of minimum wage.
Due to the nature of the proposed task, the data we release might contain false or misleading statements. In the case of annotated data, these statements are labeled as such, but this does not for the unannotated portions of the data. 
We manually pre-filtered the data to remove profanities before providing them to annotators. 
Such precautions were not taken for the unannotated portion of the dataset, which might therefore contain offensive, obscene or otherwise unconscionable items.

\section*{Acknowledgments}
The construction of the SHROOM dataset was made possible by a grant from the Oskar Öfflund Foundation.
This work is also supported by the ICT 2023 project ``Uncertainty-aware neural language models'' funded by the Academy of Finland (grant agreement  \textnumero{}~345999). 
We also thank  the CSC-IT Center for Science Ltd., for computational resources. 

The shared task logo (cf. \Cref{fig:logo}) uses the ``Retro Cool'' font from Nirmana Visual (\url{https://nirmanavisual.com/}), made available for personal / non-commercial uses.

\bibliography{anthology,custom}
\bibliographystyle{acl_natbib}

\appendix
\section{Shared consciousnesses: Overview of approaches used by SHROOM teams}
\label{adx:participants}
\begin{table*}[!h]
    \centering
    \resizebox{0.95\linewidth}{!}{
    \input{tables/teams_citations}

    }
    \caption{Participating teams and their respective works.}
    \label{tab:all_teams}
\end{table*}

In \Cref{tab:all_teams}, we provide a short overview of the various teams, the resources they utilized (models \& datasets), as well as a short description of their approach.

\section{What SHROOM makes you do: Annotation guidelines}\label{appx:annotation_guidelies}
\begin{figure*}
     \centering
   \includegraphics[max height=0.95\textheight, max width=\textwidth, trim= 0cm 15cm 0cm 0cm]{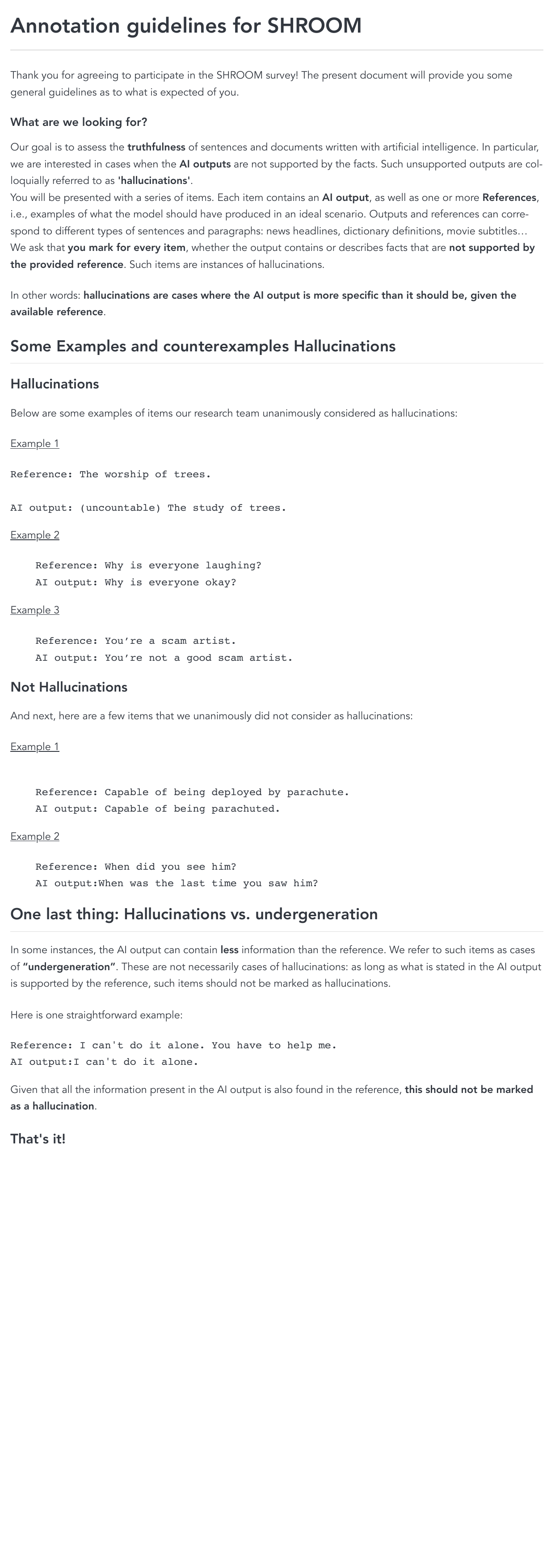}    
    \caption{Annotation guidelines.}
    \label{fig:guidelines}
\end{figure*}

In \Cref{fig:guidelines}, we provide an exact copy of the annotation guidelines given to the annotators.
These guidelines are based on five of the organizers' experience of annotating the trial set, and were provided to annotators recruited for the validation and test splits.

\end{document}

%% file: tables/rankings-agnostic.tex
 \rowcolors{2}{gray!25}{white}
 \sisetup{round-precision=3}
\begin{tabular}{>{\bf}l@{{~~}}>{\small}l@{{\quad}}S[table-format=1.3]@{{\quad}}S[table-format=2.3]}
& {\normalsize \textbf{team}} & \textbf{Acc} & $\rho$ \\ \toprule
1 & Halu-NLP \citep{no53} & 0.847333 & 0.769512 \\
2 & OPDAI \citep{no109} & 0.836000 & 0.732195 \\
3 & HIT-MI\&T Lab \citep{no269} & 0.830667 & 0.767700 \\
4 & SHROOM-INDElab \citep{no126} & 0.829333 & 0.721004 \\
5 & Alejandro Mosquera & 0.826000 & 0.709172 \\
6 & DeepPavlov \citep{no43} & 0.820667 & 0.752495 \\
7 & BruceW & 0.820667 & 0.735179 \\
8 & TU Wien \citep{no182} & 0.816667 & 0.737138 \\
9 & SmurfCat \citep{no131} & 0.814000 & 0.723473 \\
10 & HaRMoNEE \citep{no200} & 0.814000 & 0.625880 \\
11 & AMEX AI LABS & 0.812667 & 0.727682 \\
12 & Pollice Verso \citep{no231} & 0.802667 & 0.676024 \\
13 & MALTO \citep{no256} & 0.800667 & 0.681308 \\
14 & UCC-NLP & 0.795333 & 0.664244 \\
15 & Team CentreBack & 0.792000 & 0.622922 \\
16 & Atresa & 0.788000 & 0.645633 \\
17 & ustc\_xsong & 0.785333 & 0.694805 \\
18 & IRIT-Berger-Levrault \citep{no89} & 0.782667 & 0.636324 \\
19 & silk\_road & 0.780667 & 0.671599 \\
20 & AILS NTUA \citep{no234} & 0.778000 & 0.667760 \\
21 & zhuming & 0.773333 & 0.480944 \\
22 & SibNN & 0.770000 & 0.612759 \\
23 & UMUTeam \citep{no101} & 0.769333 & 0.560946 \\
24 & NootNoot \citep{no146} & 0.764667 & 0.583835 \\
25 & HalluSafe \citep{no23} & 0.763333 & 0.628870 \\
26 & Maha Bhaashya \citep{no257} & 0.748667 & 0.605255 \\
27 & DUTh \citep{no162} & 0.744000 & 0.474832 \\
28 & Compos Mentis \citep{no219} & 0.738000 & 0.595055 \\
29 & daixiang & 0.737333 & 0.582741 \\
30 & NU-RU \citep{no40} & 0.728000 & 0.595126 \\
\multicolumn{2}{c}{\small\it baseline system} & 0.6966666 & 0.40298808914014683  \\
31 & SLPL SHROOM \citep{no175} & 0.694000 & 0.423213 \\
32 & Skoltech & 0.684000 & 0.674308 \\
33 & CAISA & 0.676667 & -0.429702 \\
34 & AlphaIntellect \citep{no144} & 0.654000 & 0.294608 \\
35 & deema & 0.646000 & 0.565624 \\
36 & BrainLlama \citep{no14} & 0.624667 & 0.203636 \\
37 & Byun \citep{no42} & 0.616667 & 0.238979 \\
38 & Bolaca \citep{no253} & 0.612667 & 0.217062 \\
\multicolumn{2}{c}{\small\it most frequent guess} & 0.592667 & \\
39 & AI Blues & 0.587333 & 0.025104 \\
\multicolumn{2}{c}{\small\it random guess} & 0.5 &  \\
40 & MARiA \citep{no237} & 0.498000 & 0.025078 \\
41 & 0x.Yuan \citep{no292} & 0.460667 & 0.133720 \\
\bottomrule
\end{tabular}

%% file: tables/rankings-aware.tex
\rowcolors{2}{gray!25}{white}
\sisetup{round-precision=3}
\begin{tabular}{>{\bf}l@{{~~}}>{\small}l@{{\quad}}S[table-format=1.3]@{{\quad}}S[table-format=2.3]}
& {\normalsize \textbf{team}} & \textbf{Acc} & $\rho$ \\
\toprule
1 & HaRMoNEE \citep{no200} & 0.812667 & 0.699316 \\
2 & Halu-NLP \citep{no53} & 0.806000 & 0.714712 \\
3 & TU Wien \citep{no182} & 0.806000 & 0.707192 \\
4 & OPDAI \citep{no109} & 0.805333 & 0.680273 \\
5 & HIT-MI\&T Lab \citep{no269} & 0.804667 & 0.712325 \\
6 & SHROOM-INDElab \citep{no126} & 0.802000 & 0.655670 \\
7 & AMEX AI LABS & 0.800667 & 0.695710 \\
8 & DeepPavlov \citep{no43} & 0.799333 & 0.712693 \\
9 & silk\_road & 0.798000 & 0.686563 \\
10 & AILS NTUA \citep{no234} & 0.794667 & 0.685196 \\
11 & BruceW & 0.794000 & 0.659960 \\
12 & Team CentreBack & 0.789333 & 0.606304 \\
13 & UCC-NLP & 0.788667 & 0.644066 \\
14 & ustc\_xsong & 0.787333 & 0.657969 \\
15 & UMUTeam \citep{no101}& 0.784000 & 0.506895 \\
16 & HalluSafe \citep{no23} & 0.783333 & 0.537402 \\
17 & SmurfCat \citep{no131} & 0.782667 & 0.671200 \\
18 & Atresa & 0.782667 & 0.623914 \\
19 & IRIT-Berger-Levrault \citep{no89} & 0.781333 & 0.601276 \\
20 & Pollice Verso \citep{no231} & 0.777333 & 0.600736 \\
21 & NU-RU \citep{no40} & 0.768000 & 0.581907 \\
22 & zhuming & 0.768000 & 0.471599 \\
23 & SibNN & 0.762667 & 0.586943 \\
24 & Compos Mentis \citep{no219} & 0.756000 & 0.565698 \\
25 & DUTh \citep{no162} & 0.755333 & 0.528480 \\
\multicolumn{2}{c}{\small\it baseline system} & 0.7453333333333333 & 0.48788676869571396 \\
26 & AlphaIntellect \citep{no144} & 0.711333 & 0.426429 \\
27 & SLPL SHROOM \citep{no175} & 0.706000 & 0.426227 \\
28 & deema & 0.688000 & 0.519104 \\
29 & BrainLlama \citep{no14} & 0.671333 & 0.243766 \\
30 & daixiang & 0.649333 & 0.217670 \\
\multicolumn{2}{c}{\small\it most frequent guess} & 0.632667 & \\
31 & Bolaca \citep{no253} & 0.626000 & 0.283236 \\
32 & NootNoot \citep{no146} & 0.612667 & 0.355061 \\
33 & Byun \citep{no42} & 0.610000 & 0.234048 \\
34 & Maha Bhaashya \citep{no257} & 0.606000 & 0.209235 \\
35 & CAISA & 0.567333 & -0.100493 \\
36 & Skoltech & 0.557333 & -0.011140 \\
37 & MARiA \citep{no237} & 0.505333 & 0.008985 \\
\multicolumn{2}{c}{\small\it random guess} & 0.5 & \\
38 & octavianB \citep{no177} & 0.483333 & -0.063792 \\
\bottomrule
\end{tabular}

%% file: tables/teams_citations.tex
\rowcolors{2}{gray!25}{white}
\begin{tabular}{p{.25\linewidth} p{.4\linewidth} p{\linewidth}}
    \toprule
    \bf	Team	 \&	\bf	Paper &	\bf	Resources &	\bf	Overview		\\
    \midrule
    AI Blues	     	& \multicolumn{2}{c}{\it \small (No report)} \\
    AILS NTUA\newline   	\citet{no234}	& SHROOM datasets; Vectara model.& Fine-tuned models and voting classifier. \\
    Alejandro Mosquera 	& SHROOM datasets; COMET, Vectara, LaBSE, GPT35 and GPT4 models. & Ensemble of publicly available models. Logistic Regression was used as final scoring model. \\
    AlphaIntellect	 \newline 	\citet{no144}	& SHROOM dataset, SBERT & Fully-connected neural network classifiers with SBERT embeddings as input. \\
    AMEX AI LABS	  		& SHROOM datasets; Vectara and OpenChat models. & Ensemble of LLM (using Openchat) zero shot and few shot with Vectara cross encoder based scores. \\
    Atresa	         		& \multicolumn{2}{c}{\it \small (No report)} \\
    BrainLlama\newline   \citet{no14} & LLaMA model. & Prompt-based approach with LLaMA.	\\
    BruceW	        	& \multicolumn{2}{c}{\it \small (No report)}\\
    Byun \newline    \citet{no42}	& SHROOM dataset, data augmentation, RoBERTa & Finetuned a BERT or RoBERTa model with a softmax layer to output the probability of hallucinated text. Finetuning data is the labelled SHROOM data augmented with data points constructed by replacing words with synonyms.\\
    CAISA	         		& \multicolumn{2}{c}{\it \small (No report)}\\
    Compos Mentis \newline  \citet{no219}	& HalluEval dataset; Mistral 7B instruct model. & Ensemble of several role-based LLMs, which were either fine-tuned on hallucination data or role-based prompting.  \\
    daixiang	      			& \multicolumn{2}{c}{\it \small (No report)}\\
    deema	          		& \multicolumn{2}{c}{\it \small (No report)}\\
    DeepPavlov	 \newline  \citet{no43}	& SHROOM dataset; OpenChat, DeBERTa, RoBERTa and T5 models. & Ensemble of several pretrained Transformer-based models to get features for validation and test data of SHROOM dataset and trained a boosting-based meta-model on top. \\
    DUTh	\newline  \citet{no162}	& SHROOM, LaBSE, T5, DistilUSE & Using pre-trained LLMs and classifiers\\
    Bolaca		\newline  \citet{no253}		& SHROOM dataset, SBERT & Logistic regression and feed-forward classifier trained on SBERT embeddings\\
    HalluSafe 	\newline  \citet{no23}	&  SHROOM, labeled 3000 samples of the training data	&  Fine-tuned a DeBERTa-v3-large \\
    Halu-NLP	\newline  \citet{no53}	& SHROOM datasets; GPT, SelfCheckGPT and Vectara models. & Prompts and GroupCheckGPT. NB: due to a team name change, this team is also referred to as GroupCheckGPT by some participants. \\
    HaRMoNEE 	\newline  \citet{no200} & SHROOM, SNLI, MNLI and PAWS datasets; Vectara and GPT4 models. & Highest results obtained with zero-shot prompting in the model-aware track; pretraining on NLI and PAWS followed by finetuning on the model-agnostic track. \\
    HIT-MI\&T Lab	\newline  \citet{no269}	& SHROOM with training dataset  labeled using GPT-4; DeBERTaV3, InternLM2, SBERT, and UniEval. &   
    Fine-tune the DeBERTaV3 and InternLM2 models, and call the SBERT and UniEval models to select the optimal threshold usinf SHROOM \& syntheticaly labeled data. The system obtians the final results by combining the prediction results of each model.\\
    IRIT-Berger-Levrault	\newline  \citet{no89}	&SHROOM datasets; Sentence-t5, BGE, e5 models. & Computes the cosine similarity of sentence embeddings and classify based on an empirical threshold value. \\
    Maha Bhaashya	\newline  \citet{no257}	& DeBERTa models. & Zero shot inference, pretrained cross encoder model  \\
    MALTO	\newline  \citet{no256} & SHROOM model-agnostic dataset, DeBERTa pretrained and finetuned on MNLI, SOLAR-10.7B quantized from TheBloke (for synthetic data generation) & Encoder and classifier, fine-tuned in various ways (including with synthetic data)	\\
    MARiA	\newline  \citet{no237}	& SHROOM dataset, SBERT, bart-large-mnli, Mixtral & Three approaches: (1) Cosine similarity of SBERT embeddings between source-hypothesis and source-target pairs; (2) NLI classification using bart-large-mnli model; and (3) Mixtral prompting. Only the Mixtral results were submitted. \\
    NootNoot	\newline  \citet{no146}	& SHROOM dataset; Mixtral and RoBERTa models. &  Mixtral prompting and  RoBERTa finetuning. \\
    NU-RU	\newline  \citet{no40} & SHROOM, GPT-3.5, Sentence Transformers  &  Tried two approaches: (1) hypothesis-target cosine similarity, using a threshold value to determine whether the hypothesis is a hallucation. (2) SelfCheckGPT with a customized prompt for each NLG task, designed to assess its coherence with the provided source and target. Each prompt is iterated through the GPT-3.5 model five times, and the final label is determined by the majority response.  \\
    octavianB	\newline  \citet{no177}	& RoBERTa & Used a pretrained model (roberta-large-openai-detector) that has been trained to distinguish between text generated by LLMs and text written by humans.\\
    OPDAI	\newline  \citet{no109}	& SHROOM, Mistral-7B-Instruct-v0.2, self constructed training data & Supervised fine-tuning over synthetically constructed weakly supervised training data. \\
    Pollice Verso	\newline  \citet{no231}	& Mistral2, LLaMa2, Phi2 and Zephyr models; uses SHROOM train set for prompt optimization. & Ensembling over the output logits of prompt-based LLMs (mistral, llama etc) after automatically optimizing their prompts ("OPRO").\\
    SHROOM-INDElab	\newline  \citet{no126}	& SHROOM dataset; GPT 3.5 and GPT 4 models. & In-context learning with role-play and automatic prompt generation in a few-shot classfier, using a closed-source LLM. \\
    SibNN			&SHROOM datasets; XLM-RoBERTa model. & Fine-tunes a self-adaptive hierarchical variant of XLM-RoBERTa-XL twice: first as an embedder (in a few-shot mode), then as a binary classifier. More details at \newline \url{https://huggingface.co/bond005/xlm-roberta-xl-hallucination-detector}. \\
    silk\_road			& SHROOM datasets; Vectara model. & Fine-tunes an off-the-shelf Cross-Encoder hallucination evaluation model. \\
    Skoltech			& \multicolumn{2}{c}{\it \small (No report)}\\
    SLPL SHROOM	\newline  \citet{no175}	& SHROOM datasets; LaBSE, DeBERTa, Zephyr, Mistral and Llama2 models. & Using two LLMs to classify and explain their decision and another LLM to judge and decide based on those explanations. \\
    SmurfCat	\newline  \citet{no131}	& SHROOM (synthetically augmented), QQP and PAWS datasets; E5, T5, Vectara models. & Fine-tuning of e5-mistral-7b-instruct using synthetic data collected with LLaMA2-7B adapters trained to produce data with and without hallucinations. However, there are two other systems: one works as a voting ensemble of multiple LLMs, and another uses the Mutual Implication Score architecture. \\
    Team CentreBack			& SHROOM dataset; DeBERTa model. & Uses an off-the-shelf library (SelfCheckGPT's SelfCheckNLI function) to calculate contradiction scores on a small labeled test set and then defined a threshold for hallucination. \\
    TU Wien	\newline  \citet{no182} & SHROOM dataset; Vectara model. & Model-aware track best submissions uses a Vectara hallucination detection model finetuned on the validation set. The best model-agnostic track submission is a meta-model that utilizes linear regression and is trained on features that correspond to probabilities predicted by individual systems we implemented. 	\\
    UCC-NLP		& SHROOM dataset; GPT-3.5 and Vectara models. & Uses BertScore and GPT-3.5 to create synthetic labels and fine-tune a Vectara LLM. \\
    UMUTeam	\newline  \citet{no101}	& SHROOM dataset; TULU-DPO model. & Zero-shot approach\\
    ustc\_xsong			& \multicolumn{2}{c}{\it \small (No report)} \\
    zhuming			& \multicolumn{2}{c}{\it \small (No report)} \\
    0x.Yuan \newline \citet{no292}		& Mistral, Mixtral, LLaMA, Falcon, WizardLM and Capybara models.  & Zero-shot prompt engineering. Expects most LLMs will have different hallucination patterns, and tests whether ensembling can mitigate this. \\
    \bottomrule
\end{tabular}